\pdfoutput=1

\documentclass[letterpaper, 10 pt, conference]{ieeeconf}  % Comment this line out if you need a4paper

\IEEEoverridecommandlockouts                              % This command is only needed if 
\usepackage{url}
\usepackage{graphics} % for pdf, bitmapped graphics files
\usepackage{epsfig} % for postscript graphics files
\usepackage{mathptmx} % assumes new font selection scheme installed
\usepackage{times} % assumes new font selection scheme installed
\usepackage{amsmath} % assumes amsmath package installed
\usepackage{amssymb}  % assumes amsmath package installed
\usepackage{url}
\usepackage{subfigure}
\usepackage[T1]{fontenc}
\usepackage[vlined,ruled]{algorithm2e}
\usepackage{booktabs}
\usepackage{multirow}

\usepackage{graphicx}
\usepackage{float}
\usepackage{cite}
\usepackage{xcolor}
\usepackage{url}
\usepackage{algpseudocode}
\usepackage[export]{adjustbox}
\usepackage{balance}
\usepackage[utf8]{inputenc}
\usepackage[colorlinks,linkcolor=blue]{hyperref}
\title{\LARGE \bf
TP-TIO: A Robust Thermal-Inertial Odometry with Deep ThermalPoint		
}

\author{ Shibo Zhao$^{1}$, Peng Wang$^{2}$, Hengrui Zhang$^{1}$, Zheng Fang$^{2}$, Sebastian Scherer$^{1}$
	\thanks{
	$^{1}$ Shibo Zhao, Hengrui Zhang, Sebastian Scherer are with Robotics Institute, Carnegie Mellon University, USA, {\tt\scriptsize \{shiboz,hengruiz,basti\}@andrew.cmu.edu}}
	\thanks{$^{2}$ Peng Wang and Zheng Fang are with Faculty of Robot Science and Engineering, Northeastern University, China 
		{\tt\scriptsize \{pengwang,fangzheng\}@mail.neu.edu.cn}}}

\begin{document}
	\maketitle
	\thispagestyle{empty}
	\pagestyle{empty}

% 	To extend applications of the robot in GPS-denied and visually degraded environments such as darkness and smoke, thermal Odometry has been an attraction in the robotic community, and the approaches that most people came up with are based on classical feature extractors on re-scaling thermal images. However, it is not suitable for re-scaling thermal images which have heavy photometric changes and the low signal-to-noise ratio, especially fixed pattern noise. In this paper, we propose to leverage deep thermal point detection to overcome the limitations of feature-based thermal odometry. To this end, we combine thermal point detection with and visual-inertial bundle adjustment to provide a robust odometry in severely visually degraded environments. For the thermal point detection, we design a lightweight network and propose some improvements including network architecture and training methods to improve robustness and real-time performance. Finally, an extensive set of experiments is conducted to thoroughly evaluate the proposed algorithm from ThermalPoint to odometry in all kinds of environments including obscurants-filled underground mines and dense smoke-filled indoor environments, demonstrating superior performances of our methods. 
	%%%%%%%%%%%%%%%%%%%%%%%%%%%%%%%%%%%%%%%%%%%%%%%%%%%%%%%%%%%%%%%%%%%%%%%%%%%%%%%%
	\begin{abstract}
 To achieve robust motion estimation in visually degraded environments, thermal odometry has been an attraction in the robotics community. However, most thermal odometry methods are purely based on classical feature extractors, which is difficult to establish robust correspondences in successive frames due to sudden photometric changes and large thermal noise. To solve this problem, we propose ThermalPoint, a lightweight feature detection network specifically tailored for producing keypoints on thermal images, providing notable anti-noise improvements compared with other state-of-the-art methods. After that, we combine ThermalPoint with a novel radiometric feature tracking method, which directly makes use of full radiometric data and establishes reliable correspondences between sequential frames. Finally, taking advantage of an optimization-based visual-inertial framework, a deep feature-based thermal-inertial odometry (TP-TIO) framework is proposed and evaluated thoroughly in various visually degraded environments. Experiments show that our method outperforms state-of-the-art visual and laser odometry methods in smoke-filled environments and achieves competitive accuracy in normal environments.
		
		%indoor environments, as well as in dense somke-filled environments.
		%It shows that our method can achieve accurate and robust pose estimation compared with other state-of-the-art visual odometry method and outperforms the state-of-the-art laser odometry inside a smoke-filled setting under severely visually degraded conditions.
		
	\end{abstract}
	
	%%%%%%%%%%%%%%%%%%%%%%%%%%%%%%%%%%%%%%%%%%%%%%%%%%%%%%%%%%%%%%%%%%%%%%%%%%%%%%%%
\section{INTRODUCTION}
\label{sec:introduction}

Robust and accurate state estimation for micro aerial vehicles (MAVs) is of crucial importance as these versatile robots are taking the place of people to fulfill complex and dangerous missions, such as industrial inspection, remote sensing, search and rescue.
Due to their flexibility and ability to minimize risks to humans, there is an increased demand for their ability to achieve perception and localization safely and reliably in more challenging environments, such as fire scenes, subterranean settings, as well as GPS-denied environments. To localize in such environments, visible-light camera sensors seem to be a suitable choice for MAVs because of low power consumption, light weight and affordable price. A number of promising visual odometry and visual-inertial odometry methods have been proposed in recent years. Nevertheless, due to the poorly illuminated conditions and airborne obscurants conditions such as dust, fog and smoke in real world, the data of visible-light cameras can significantly degrade, which makes them unreliable for motion estimation.
%These environments are not only GPS-denied but also poorly illuminated. The presence of airborne visual obscurants such as dust, fog and smoke also makes such environments more challenging.
%To navigate in such environments, visible-light camera sensors seem to be a suitable choice for MAVs because of low power consumption, light weight and affordable price. A number of promising visual odometry and visual-inertial odometry methods have been proposed in recent years. Nevertheless, due to the poorly illuminated conditions and airborne obscurants conditions, the data of visible-light cameras can significantly degrade, which makes them unreliable for motion estimation.
\begin{figure}[htbp]
	\centering
	\includegraphics[width=1.0\columnwidth]{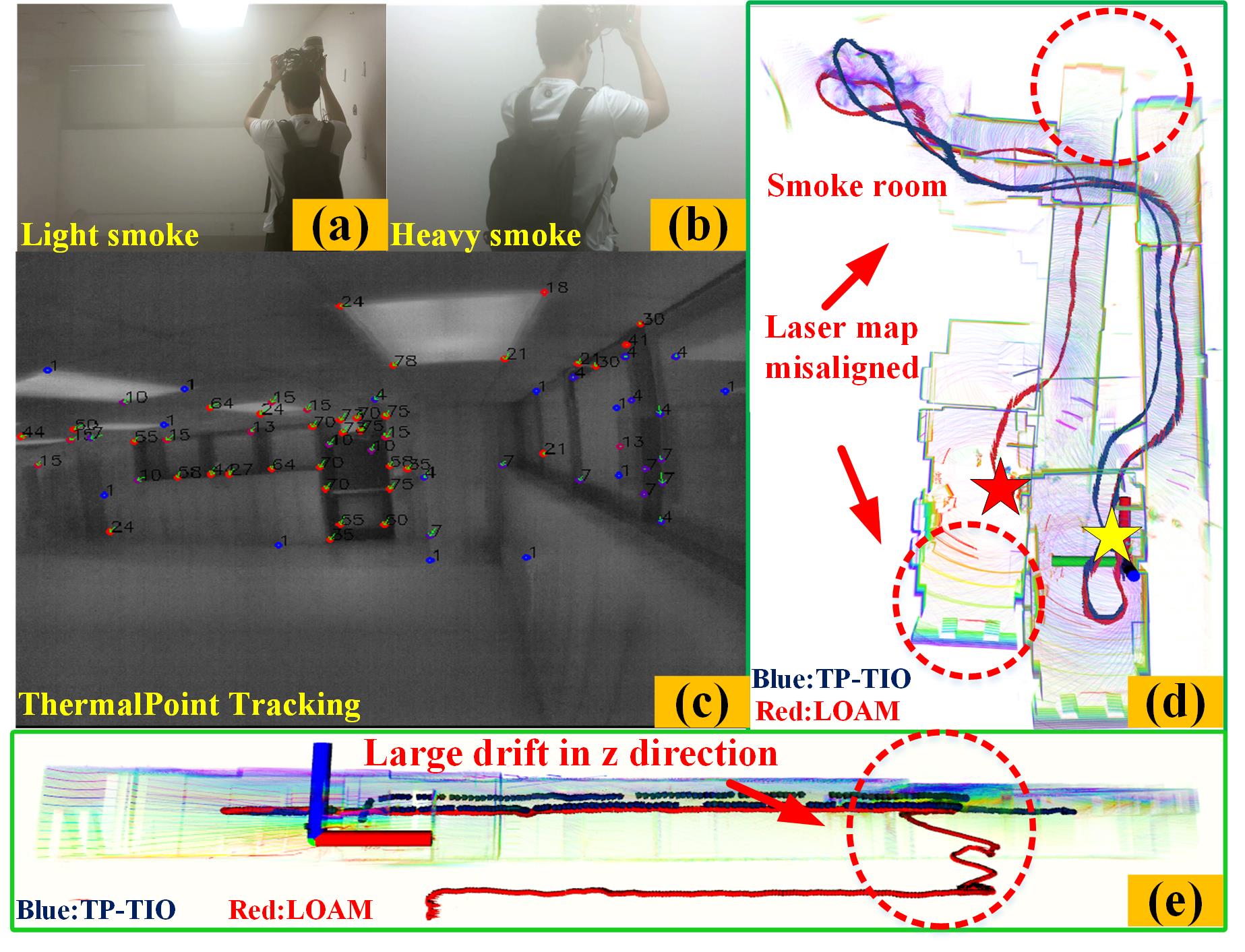}
	\caption{The perception abilities comparison of different sensors in the same smoke-filled environments. (a) and (b) show the real scene of light-smoke and dense-smoke from a visible-light camera. (c) shows the ThermalPoint tracking results from a thermal camera. Red point means that the feature has been tracked more than ten times. Otherwise, it will be marked as a blue point. (d) shows the trajectory comparison of TP-TIO (blue) and LOAM (red). The red and yellow stars represent the final position of LOAM and TP-TIO, respectively. It can be seen that our method is closer to the origin (xyz axes) while LOAM has a large pose estimation drift and point cloud misalignment. (e) shows the trajectory comparison in the z-direction. It can be seen that our method has a small drift in the z-direction.}
	\label{fig:first_page}
	\vspace{-3mm}
\end{figure} 

Compared with visible-light cameras, thermal cameras are robust to illumination variations and obscurants like dust and fog. Thermal cameras, especially the thermal cameras operating in the long wave infrared spectrum (LWIR), have better spectral penetration properties in the presence of fog, dust, and smoke (shown in Fig.\ref{fig:first_page}). Therefore, thermal odometry is an ideal choice to achieve robust pose estimation in visually degraded environments.
In spite of this, the use cases of the thermal camera are largely limited to perception and localization. Since the thermal camera captures 16-bit temperature radiation images which have low contrast, it is difficult to use handcrafted feature methods to extract salient features. To solve this problem, most of the previous methods \cite{borges2016practical,mouats2014multispectral} re-scale 16-bit thermal images to 8-bit to enhance the contrast of thermal images and improve the performance of feature matching. However, re-scaling thermal image data will result in loss of information, amplifying the effect of noise and image saturation \cite{khattak2019keyframe}. Moreover, thermal images have fixed-pattern noise and requires a  Non-Uniformity Correction (NUC) process \cite{borges2016practical} to remove noise, which  periodically suspends the images for up to 500 milliseconds. This kind of image interruption can easily cause the failure of feature tracking and drift of state estimation. 

In the last decade, due to the ability to extract high-level features automatically, deep neural architectures have been proved advantageous for feature extraction in various data types including RGB images and point clouds.
% Therefore, we believe that, with  proper and sufficient training data, deep learning based method has potential to overcome the noise  of thermal images and extract more robust features compared with traditional methods. 
However, the shortcomings of deep learning based feature extraction networks are also obvious, such as high computation cost and poor real-time performance. 

Motivated by the discussion above, we propose a novel deep thermal-inertial odometry (TP-TIO).
% The proposed method can robustly extract and track features on thermal images as well as estimate the ego-motion of sensors in various difficult environments with smoke and darkness. 
The main contributions of our work are as follows:

\begin{itemize}
	\item 	
	    We propose the first tightly coupled deep thermal-inertial odometry algorithm (TP-TIO), which utilizes both a lightweight learning-based feature detection network and an optimization-based visual-inertial framework.
	
	\item We propose the ThermalPoint, a CNN-based feature detection network specifically tailored for producing keypoints on thermal images, that provides notable improvements in anti-noise compared with  other state-of-art learning-based and handcrafted feature extraction methods.
	
	\item We propose an IMU-aided radiometric feature tracking method, which directly makes use of full radiometric data from a infrared camera to establish robust correspondences between consecutive frames.
	
	\item We conducted an extensive set of experiments to thoroughly evaluate our framework from thermal feature tracking to pose estimation in all kinds of environments including obscurant-filled underground mines and dense smoke-filled indoor environments.
	 
\end{itemize}
The rest of the paper is organized as follows: Section \ref{sec:relatedwork}
reviews various feature extracton methods and thermal-inertial odometry methods. Section \ref{sec:algorithm} illustrates the details of the proposed
method. Quantitative and qualitative evaluations from ThermalPoint to state estimation are presented in Section \ref{sec:experiment}. Section \ref{sec:conclusion} concludes the paper.
%%%%%%%%%%%%%%%%%%%%%%%%%%%%%%%%%%%%%%%%%%%%%%%%%%%%%%%%%%%%%%%%%%%%%%%%%%%%%%%%     
\section {Related Work}
\label{sec:relatedwork}
In this section, we start by reviewing existing feature extractor methods, which are essential for vision-based and thermal-based motion estimation, then we review various thermal odometry or thermal SLAM methods.
\subsection{ Feature Extractor on Thermal Images}
In decades, various handcrafted feature point detection and descriptor algorithms have been proposed and successfully utilized in many visual odometry and SLAM framework. However, since the classical feature extractors are based on gradient information of the images, they cannot work well when images have severe transforms (e.g. viewpoint, blurring, noise). To solve this problem, recent works\cite{superpoint, LIFT,gcnv2} leverage deep learning methods to generate more robust interest points and descriptors. Among these methods, LIFT \cite{LIFT} performs both keypoint detection and descriptor extraction based on patch methods and requires a supervision strategy. SuperPoint \cite{superpoint} proposes a homographic adaptation strategy and predicts the keypoints and descriptors from an unsupervised approach. GCNv2 \cite{gcnv2} adopts a binary descriptor with good computational efficiency. However, all of these methods are designed for standard visible band images and they are difficult to achieve good performance on thermal imagery.

\subsection{Thermal Odometry}
Robust and accurate state estimation from a thermal camera remains a challenging problem. In recent years, there are a number of efforts that have been made for the thermal odometry. 
% Mouats et al\cite{mouats2015thermal} proposed a thermal stereo odometry using Fast-Hessian feature detector.
To improve the robustness of thermal odometry, many works fuse thermal images with other modalities like RGB images\cite{mouats2014multispectral,poujol2016visible,dai2019multi} or laser scanner\cite{shin2019sparse}.
% Chen et al. proposed a RGB-T SLAM applying ORB features on both thermal and visible images.
% Papachristos et al proposed a thermal-inertial odometry employed fast features. 
Since the above methods use traditional handcrafted feature detection methods that mainly rely on the gradient of images, it is difficult for these methods to overcome the effect of large noise. To solve this problem, we propose the "ThermalPoint", which adds a noise reduction module during the training process and effectively overcomes the noise of thermal images. Khattak et al.\cite{khattak2019keyframe} developed a keyframe-based thermal inertial odometry and proved the effectiveness of the direct method. However, since the direct methods heavily rely on image quality and accurate initial pose from the back-end module, it is particularly challenging for these methods to provide robust motion estimation when the environments have a scarcity of thermal gradients. In contrast, since our feature tracking method doesn't rely on the feedback of the back-end module, it can handle temporary feature tracking failure in environments with no salient heat sources.
% lack of a public implementation does not allow us to perform a comparison of accuracy, robustness.
Muhamad\cite{saputra2020deeptio} proposed an end-to-end thermal inertial odometry. However, this method can only perform well with a slow frame rate. To achieve motion estimation at a high frequency, our method uses ThermalPoint in a shallow architecture and radiometric feature tracking method to establish reliable correspondences instead of matching descriptors. This greatly improves real-time performance and reduces computation.
 
In summary, it is relatively difficult to find a real-time and robust deep feature-based thermal-inertial odometry under severely visually degraded conditions in the existing literature. This paper aims to solve this problem, providing a reliable sparse feature-based state estimation method in the thermal image domain.
	%%%%%%%%%%TODO:
%%%%%%%%%%TODO:
\section{Thermal-Inertial Odometry} \label{sec:algorithm}

%\begin{figure}[htbp]
%	\centering
%	\includegraphics[width=0.9\columnwidth]{result/pipeline.jpg}
%	\caption{ Overview of Laser-Inertial Odommetry and Mapping Method}
%	\label{fig:algorithm}
%	\vspace{-3mm}
%\end{figure} 

\begin{figure*}
\centering
\includegraphics[width=2.0\columnwidth]{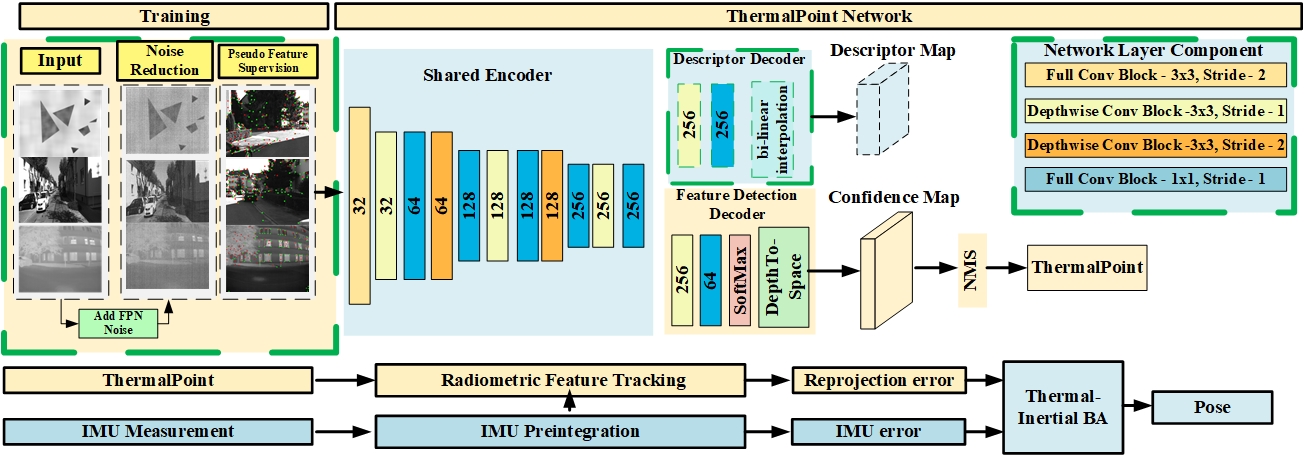}
\caption{Overview of deep thermal-inertial odometry. The modules marked as yellow are our contributions to this work. To improve real-time performance, the modules with green dashed squares will only participate in the training process and be excluded in a real application.}
\vspace{-3mm}
\label{fig:algorithm}
\end{figure*}

In this section, we will introduce the pipeline of robust Thermal-Inertial Odometry with deep "ThermalPoint" (TP-TIO) as shown in Fig.\ref{fig:algorithm}, which allows robust motion estimation in severely visually degraded environments. It is composed of three sequential modules, namely ThermalPoint network, IMU-aided radiometric feature tracking and thermal-inertial bundle adjustment.
\subsection{ThermalPoint Network}

\subsubsection{Network Architecture}

To detect features robustly on noisy thermal images, we designed a fully convolutional neural network architecture called ThermalPoint shown in Fig.\ref{fig:algorithm}. Inspired by the famous SuperPoint \cite{superpoint} network, this network has a shared encoder which abstracts high-level features and reduces the dimensionality from the input image. After that, the architecture is divided into two task decoders, feature detection decoder and descriptor decoder, respectively. To improve real-time performance,
the descriptor decoder is only used to improve the repeatability\cite{superpoint} of feature detection during the training process and is removed in the inference process. Instead, we use KLT-based feature tracking method to establish feature correspondences. More details can be found in Section \ref{sec:tracking}.

\subsubsection{Shared Encoder}
% 说明vgg-net 的优缺点
Considering the low latency and computation cost for feature detection on embedded systems, our shared encoder is built from an efficient network, MobileNet\cite{howard2017mobilenets}, instead of the  VGG-style shared encoder used by SuperPoint. The specific structure of our shared encoder can be found in Fig.\ref{fig:algorithm}.

%A number of impressive feature detection networks use a VGG-style as the base feature extractor \cite{simonyan2014very} because of the expressive power and high accuracy. In spite of this, such networks also suffer from unnecessary complexity and latency, which is not suitable for real-time applications running on embedded systems.
%
%% 说明mobilene的优点
%To achieve low latency for feature detection, our shared encoder is built from an efficient network, MobileNet \cite{howard2017mobilenets}, which makes use of depthwise separable convolution and $1\times1$ pointwise convolution to replace standard convolution. 
% 说明我们的网络的架构
%Our shared encoder starts with a full convolution layer to improve the accuracy of feature extraction, and contains five depthwise convolution layers, and five 1x1 pointwise convolution layers. All these layers are followed by a batchnorm \cite{ioffe2015batch} and ReLU nonlinearity. Down sampling is handled with strided convolution in the depthwise convolution. The shared encoder maps the input image $I\in \mathbb{R}^{H\times W}$ to a temporary  high-level low-resolution feature map $B \in \mathbb{R}^{H_c \times W_c \times F}$ where each pixel corresponds to a non-overlapping $8\times8$ pixel region. 

\subsubsection{Feature Detection Decoder And Descriptor Decoder}

Inspired by SuperPoint \cite{superpoint}, the feature detection decoder computes a feature map $X\in \mathbb{R}^{H_c\times W_c \times 64}$.
% by filtering  previous feature map $B$. 
 After that, it will output the final confidence heatmap $C$ sized $H \times W$ by performing a channel-wise softmax and DepthToSpace. Each pixel in $X$ corresponds to a probability, measuring a likelihood to be the feature point in the $8\times 8$ cell. For the descriptor part, ThermalPoint is similar to the SuperPoint. Here are some differences we have improved between ThermalPoint and SuperPoint \cite{superpoint}.

\begin{itemize}
	\item 	
    To improve the repeatability\cite{superpoint} of feature points, we replace multiple cross-entropy loss used by SuperPoint with an independent logistic loss (Eq.(\ref{eq:detector loss})). The reason is that multiple cross-entropy loss only allows one feature extracted in the cell, while independent logistic loss encodes the probability for every pixel.
	\item To save computation time during training, our decoder performs lower computational bilinear interpolation $Y$ to get dense descriptor map $D$, instead of bi-cubic interpolation of the SuperPoint.
\end{itemize}

%For SuperPoint network, since it uses the multiple cross-entropy as the feature point detection loss, it can only extract one feature point per cell. However, for the case that a cell may contain multiple feature points, multiple cross-entropy may not be an ideal choice. To address this problem, we replace multiple cross-entropy with an independent logistic loss (Eq.(\ref{eq:detector loss})) which encodes the probability for every pixel. Through this process, we can extract more potential features in the thermal images and improve the repeatability of feature points. For the descriptor part, similar to the SuperPoint‘s descriptor decoder, our descriptor decoder predicts a feature map $Y \in \mathbb{R}^{ H_c \times W_c\times D}$ by filtering $B$ and channel-wise L2 normalization.
%To save computation time during training, our decoder  performs lower computational bilinear interpolation $Y$ to get dense descriptor map $D$, instead of bi-cubic interpolation of the SuperPoint.

\subsubsection{Loss Function}
Similar to the SuperPoint's training process, we optimize both the feature point detector loss and the descriptor loss.
For the descriptor loss, we directly employ the SuperPoint's loss. The feature point detector loss function $L_{dec}(I)$ is different, where we use a binary logistic loss over each pixel of $C$. Given the feature point label $L(I) \in \mathbb{R}^{H \times W}$for each image $I$, the loss can be defined as: 
\begin{equation}
\begin{aligned}
\label{eq:detector loss}
L_{dec}(I) = \frac{1}{HW} \sum_{i=1,j=1}^{H,W} -log(C(i,j))L(i,j) - \\
log(1 - C(i,j))(1-L(i,j))
\end{aligned}
\end{equation}

\subsection{Training Details}
%Firstly, we train a base feature detector on the synthetic dataset by only minimizing the detector loss, then we refine it on the KITTI \cite{geiger2013vision} dataset. After getting a trained feature detector, we fine-tune the detector on a more difficult mixture of the KITTI dataset and a self-recorded infrared dataset by jointly minimizing detector loss and descriptor loss. We use a batch size of 32, an Adam Optimizer with a learning rate of 10−3 and a decay factor of 0.5 after 20 epochs in training process.
%In the reminder of this section, we describe dataset augmentation \ref{Data_augmentation} and the approach of producing the feature point labels \ref{multiple_feature_point_labels}.
\subsubsection{Noise Reduction} \label{Data_augmentation}

\begin{figure}[htbp]
	\centering
	\includegraphics[width=.4\textwidth ]{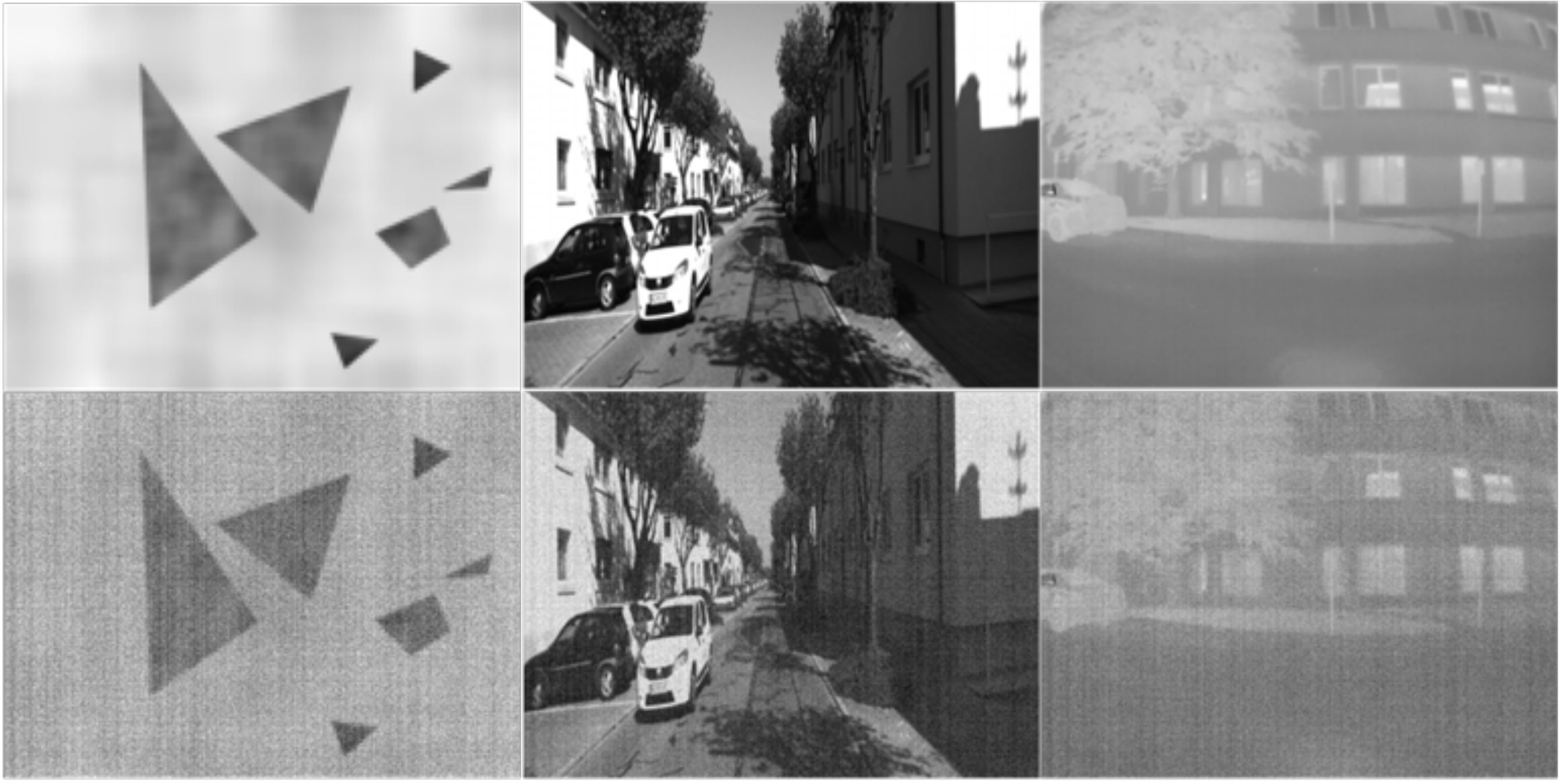} 
	\caption{Samples of applying FPN on datasets. The first row shows original noise-free images from synthetic dataset, KITTI dataset and thermal dataset. The bottom row shows corresponding non-uniform images after adding FPN noise.}
	\label{fig:fpn}
\end{figure}

%\begin{figure}[htbp]
%	\centering
%	\includegraphics[width=.4\textwidth ]{experiment_result/imagingpipeline.png} 
%	\caption{the imaging pipeline and associate noise processes}
%	\label{fig:imagepipeline}
%\end{figure}
 Applying various noise models to training datasets is a common technique to improve the model's anti-noise performance. Compared with the noise presented in visual images, thermal images have more complex noises, especially Fixed Pattern Noise (FPN) which is also called non-uniform noise. To overcome the effects of FPN on feature extraction, we need to add FPN noise to the training datasets (Fig.\ref{fig:fpn}). Given the non-uniform image $\tilde{I}(i,j)$ and the noise-free image $I(i, j)$, the relation of them can be simply described as the below equation
\begin{equation}
\label{eq:nonuniform}
\tilde{I}(i,j) = I(i,j) + n(i,j)
\end{equation} 
where $n(i,j)$ is the non-uniform noise.

To obtain non-uniform noise, we employ two simple and effective methods: The first is facing a thermal camera toward a scene with the same material like a concrete wall. In terms of their heat capacity, the non-uniform noise will be recorded due to the scarcity of thermal gradients. The second is to simulate column FPN. In this paper, we use both the above two methods to create a non-uniform noise dataset which contains 1000 images. During the training process, we randomly select an image from the non-uniform noise dataset and add it to a noise-free image based on Eq.(\ref{eq:nonuniform}) (See Fig.\ref{fig:fpn}). Meanwhile, we also use standard data augmentation tricks including contrast, hue, brightness to improve the network’s robustness against photometric changes.
\subsubsection{Pseudo Feature Supervision} \label{multiple_feature_point_labels}
\begin{figure}[htbp]
	\centering
	\includegraphics[width=.48\textwidth ]{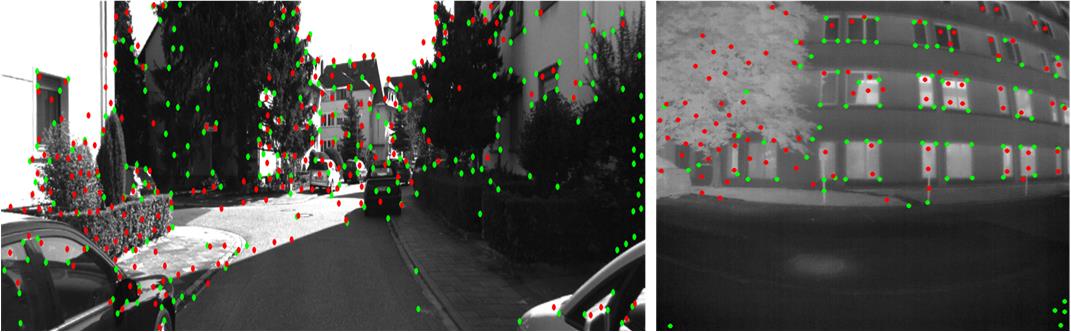} 
	\caption{Pseudo feature supervision training on RGB (left) and Thermal dataset (right), respectively. The green and red points represent keypoints detected by our base feature detector and SIFT \cite{SIFT} method, respectively. Our feature detector mainly detects some "conner" points, while the SIFT method can detect some "salient gradient" points. Both  keypoints will be used as labels during the training process.}
	\label{fig:label}
\end{figure}

SuperPoint \cite{superpoint} shows great benefits of using anchors to guide its training process. Although anchors provide a stabler training process, they also prevent the model from identifying new keypoints when no anchor is in the proximity. To solve this problem, we use SIFT \cite{SIFT} to generate some extra pseudo keypoint labels and supervise the training process. By combining the labels from handcrafted features and learned CNN filters, the repeatability of ThermalPoint is improved, as shown in Table\ref{tab: feature detectors repeatability}. The specific detail can be found in Fig.{\ref{fig:label}}.

	%%%%%%%%%%%%%%%%%%%%%%%%%%%%%%%%%%%%%%%%%%%%%%%%%%%%%%%%%%%%%%%%%%%%%%%%%%%%%%%%  
	\subsection{IMU-Aided Radiometric Feature Tracking} 
	\label{sec:tracking}
	
After detecting a sparse set of keypoints through our thermal point network, we need to track the motion and build the correspondence of these points between frames. Since most of deep learning-based feature detection networks \cite{LIFT, superpoint} create dense descriptor map to find correspondences of features, it is difficult to meet our real-time requirements. Inspired by the work\cite{khattak2019keyframe}, instead of using the descriptor method, we use full radiometric based KLT method and IMU measurement to achieve robust feature tracking. To overcome the sudden photometric changes, we combine the inverse-compositional approach \cite{inversecomposition} with a patch dissimilarity norm, making our feature tracking invariant to intensity scaling.
	
    Given a set of patches in a reference image, we minimize the radiometric error between corresponding patches in two consecutive frames as estimating the transform $\mathbf{T} \in \mathrm{SE}(2)$. For a single residual in a patch, it can be defined as
	
	\begin{equation}
	r_{i}(\boldsymbol{\xi})=\frac{I_{t+1}\left(\mathbf{T} \mathbf{x}_{i}\right)}{\overline{I_{t+1}}}-\frac{I_{t}\left(\mathbf{x}_{i}\right)}{\overline{I_{t}}} \quad \forall  \mathbf{x}_{i} \in \Omega
	\end{equation}
	
	Where $I_{t}(\mathbf{x})$ is the thermal value of the 16-bit image $t$ at pixel location $x$. The mean thermal value if the patch in image t is $\overline{I_{t}}$. $ \Omega$ defines the set of image coordinates in the patch. 
	
% 	Finally, we calculate the radiometric error of all patches in each frame.
% 	Since we calculate the radiometric error of all patches in each frame, the final radiometric patch tracking problem can be described

% 	\begin{equation}
% 	\mathbf{e}_{\text {radio }}=\sum_{i \in P} \sum_{p \in N_{i}}\left\|r_{i}(\boldsymbol{\xi})\right\|_{2}
% 	\end{equation}
	
% 	Here, $\mathbf{e}_{\text {radio }}$ is the squared sum of radiometric errors for a set of tracked patches $(P)$ calculated over the neighborhood
% 	$(Ni)$ of each feature point.
	
	To improve the robustness for large displacements and real-time performance, we perform our patch alignment over an image scale-space pyramid from coarsest level to finest level, where the alignment results from the upper level provide a prior for the lower level. Furthermore, we use the IMU preintegration \cite{VINS} to provide transformation prior for the alignment process at the coarsest level. These two steps greatly improve the solution convergence speed and improve the robustness of the tracking.
	
	\subsection{Thermal-Inertial Bundle Adjustment} 
	
	To estimate the motion of the sensor, an optimization is performed that jointly minimizes visual reprojection error terms of the tracked features and IMU error terms based on IMU preintegration measurements within a sliding window.
	
	\subsubsection{Reprojection error} The first residual term which we use for pose estimation is the reprojection error based on thermal images. Thermal reprojection error can be described as
	\begin{equation}
	\mathbf{e}_{\text {reproj }}=p^{{c_{j}}}-K\left(T_{{b}}^{c} T_{{w}}^{b} l^{{w_{i}}}\right)
	\end{equation}
	Where $p^{{c_{j}}}$ is the point in the target frame $j$ at camera image coordinates $c$. $l^{{w}}$ is the corresponding 3D landmark location in  the reference frame $i$ at world coordinates $w$. $T_{{w}}^{b} $ is the transformation from world frame $w$ to IMU body frame $b$. $T_{{b}}^{c}$ is the transformation from camera frame $c$ to IMU body frame $b$.
	
	\subsubsection{IMU error} The second residual term is derived from IMU data based on IMU preintegration measurements. To achieve high frequency and robust pose estimation, we preintegrate several consecutive IMU measurements and add the IMU factor between frame $i$ and frame $j$. Similar to \cite{VINS}, the corresponding preintegration between IMU body frame $b_{k}$ and $b_{k+1}$ is $\Delta \mathbf{r}=(\hat{\boldsymbol{\alpha}}_{b_{k+1}}^{b_{k}}, \hat{\boldsymbol{\beta}}_{b_{k+1}}^{b_{k}}, \hat{{\gamma}}_{b_{k+1}}^{b_{k}})$. Therefore, the error term for inertial measurement within two consecutive frames can be defined as:
	
	\begin{equation}
	\mathbf{e}_{\text {imu }}=\left[\begin{array}{c}
	{\delta \boldsymbol{\alpha}_{b_{k+1}}^{b_{k}}} \\
	{\delta \boldsymbol{\beta}_{b_{k+1}}^{b_{k}}} \\
	{\delta \boldsymbol{\theta}_{b_{k+1}}^{b_{k}}} \\
	{\delta \mathbf{b}_{a}} \\
	{\delta \mathbf{b}_{g}}
	\end{array}\right]
	\end{equation}
	$=\left[\begin{array}{c}
	{\mathbf{R}_{w}^{b_{k}}\left(\mathbf{p}_{b_{k+1}}^{w}-\mathbf{p}_{b_{k}}^{w}+\frac{1}{2} \mathbf{g}^{w} \Delta t_{k}^{2}-\mathbf{v}_{b_{k}}^{w} \Delta t_{k}\right)-\hat{\boldsymbol{\alpha}}_{b_{k+1}}^{b_{k}}} \\
	{\mathbf{R}_{w}^{b_{k}}\left(\mathbf{v}_{b_{k+1}}^{w}+\mathbf{g}^{w} \Delta t_{k}-\mathbf{v}_{b_{k}}^{w}\right)-\hat{\boldsymbol{\beta}}_{b_{k+1}}^{b_{k}}} \\
	{2\left[\mathbf{q}_{b_{k}}^{w_{k}^{-1}} \otimes \mathbf{q}_{b_{k+1}}^{w_{w}} \otimes\left(\hat{\gamma}_{b_{k+1}}^{b_{k}}\right)^{-1}\right]_{x y z}} \\
	{\mathbf{b}_{a b_{k+1}}-\mathbf{b}_{a b_{k}}} \\
	{\mathbf{b}_{w b_{k+1}}-\mathbf{b}_{w b_{k}}}
	\end{array}\right]
	\\
	\\
	$
	Where $(\cdot)^{w}$ and $(\cdot)^{b}$ are world frame and IMU body frame respectively. $[\cdot]_{x y z}$ extracts the vector part of quaternions $\mathbf{q} $. Both rotation matrices $\mathbf{R}$ and Hamilton quaternions $\mathbf{q} $ are used to represent rotation. $\mathbf{p}_{b_{k}}^{w}$, $\mathbf{v}_{b_{k}}^{w}$ and $\mathbf{q}_{b_{k}}^{w}$ are translation, velocity and rotation from the body frame to the world frame while taking the $k$th image. $\mathbf{g}^{w}$ is the gravity vector in the world frame.
	$\mathbf{b}_{a}$ and $\mathbf{b}_{w}$ are accelerometer bias and gyroscope bias respectively. For more specific details about the preintegration theory, we refer the reader to \cite{VINS}.
	
	\subsubsection{Joint Optimization}
	
Since we have obtained the reprojection and inertial residual equations, for each new frame, we minimize the cost function that consists of reprojection residuals, IMU residuals and a marginalization prior $\mathbf{e}_{m}$.
	
	\begin{equation}
	J=\sum_{ i \in \mathbf{obs}(i)}\mathbf{e}_{\text {reproj }}^{\mathbf{T}} W_{\text {reproj}}^{-1} \mathbf{e}_{\text {reproj }}^{\mathbf{}}+\sum_{(a, b) \in \mathbf{C}}\mathbf{e}_{\text {imu}}^{\mathbf{T}} W_{imu}^{-1} \mathbf{e}_{imu}^{\mathbf{}}
	+\mathbf{e}_{m}
	\end{equation}
	
	$obs(i)$ represents a set that contains the ThermalPoint $i$ that is tracked and observed by other frames. The set $\mathbf{C}$ contains pairs of frames $(a, b)$ connected by IMU factors. $W_{\text {reproj}}$ and $W_{\text {imu}} $ are co-variance matrices for visual and IMU measurements. Through optimizing this cost function, we can obtain the sensor pose.
	% using Ceres solver \cite{ceres-solver}
	
	\begin{figure*}[hbt]
		\centering
		\includegraphics[width=1.95 \columnwidth, height=0.5 \columnwidth]
		{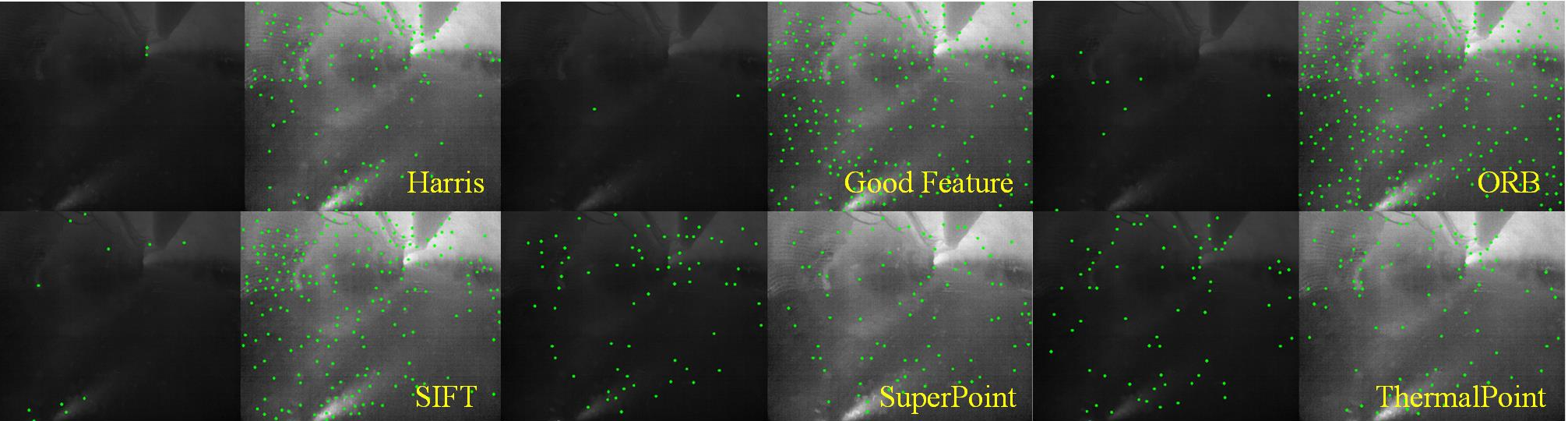} 
		\caption{The anti-photometric comparison of different feature detectors with sudden photometric changes.}
		\label{fig:feaure_illumi}
	\end{figure*}
	
	%\subsection{Marginalization}  
	%
	%In order to reduce the computational complexity of our optimization-based thermal-inertial odometry, older frames and their associated observations need to be periodically removed through marginalization process. Our marginalization strategies involves two cases. In first case, if the last frame is a keyframe, it will stay in the window, and the oldest frame will be marginalized out with its corresponding observations. However, it should be noted that the last frame can also be a nonkeyframe. Hence as a second case, all the visual measurements of the last frame will be removed but keep IMU measurements associated with this nonkeyframe. It's worth noting that  our marginalization process doesn't remove all measurements for nonkeyframe to maintain the sparsity of the system. After that, we use Schur complement [39] to finish the whole marginalization process.
	
	\section{Experiments} 
	\label{sec:experiment}
	To thoroughly evaluate the proposed TP-TIO framework and validate the performance of ThermalPoint, a variety of experiments were performed including underground mine, indoor laboratory and dense smoke-filled environments.
	\begin{itemize}
		\item For the thermal feature detection and tracking, we mainly compare the robustness and real-time performance of ThermalPoint detection and tracking with other state-of-the-art methods including Harris \cite{harris}, Good feature \cite{goodfeature}, ORB \cite{ORB}, SIFT \cite{SIFT} and SuperPoint \cite{superpoint}.
		\item For the state estimation, we mainly compare the accuracy and robustness of TP-TIO with other state-of-the-art odometry methods such as VINS \cite{VINS}, OKVIS \cite{OKVIS}, R-VIO \cite{R-VIO} and DSO \cite{DSO}.
	\end{itemize}	 
	The software runs on Jetson AGX Xavier computer with a Linux system running Robot Operating System (ROS).
	Our experiment video is available at: \textcolor{blue}{\href{https://youtu.be/aa4whgmYTqY}{https://youtu.be/aa4whgmYTqY}}

	%\begin{figure*}[h]
	%	
	%	\subfigure[OURS and VINS]{
	%		\label{ours2}
	%		\includegraphics[width=0.6\columnwidth]{experiment_result/odometry_comparison/ours_vins.jpg}
	%	}
	%	\hspace{0.000001\columnwidth}
	%	\subfigure[OKVIS and R-VIO]{
	%		\label{suma2}
	%		\includegraphics[width=0.61\columnwidth]{experiment_result/odometry_comparison/okvis_rvio.jpg}
	%	}
	%	\hspace{0.000001\columnwidth}
	%	\subfigure[Pose estimation in Z direction]{
	%		\label{loam2}
	%		\includegraphics[width=0.6\columnwidth,height=5.3cm]{experiment_result/odometry_comparison/z_direction.jpg}
	%	}
	%	
	%	\caption{The comparison of pose estimation after applying Ours (a1,c1), VINS(a2,c2) , OKVIS(b1,c3) and R-VIO(b2,c4) algorithms respectively in indoor environments.  }
	%	\label{fig:accuracy}
	%\end{figure*}

	% 特征点实验
	
	\subsection{Robustness Comparison of ThermalPoint Detection}
	
	To verify the robustness of ThermalPoint detection, we evaluate anti-photometric performance and anti-noise performance.
	For the anti-photometric test, we use a public underground mine dataset containing severe thermal photometric changes. For the anti-noise test, we use our indoor smoke dataset presenting the challenges of large noise and dense smoke. ThermalPoint is evaluated against other famous state-of-the-art feature detection methods such as Harris, GoodFeature, ORB, SIFT and SuperPoint. We select a suitable fixed threshold for each feature detector to ensure most features can be extracted on this dataset. For this test, we compute a maximum of 400 points for all systems at a 512x640 resolution in each image. The Non-Maximum Suppression(NMS) of all systems is 8.

	%
	%Furthermore, we evaluate the real-time performance of our feature detection and tracking method compared with other related  traditional and deep-learning based methods.
	
	\subsubsection{ Anti-photometric and Anti-noise Comparison} \label{Anti-photometric comparison}
\begin{figure}[h]
		\centering
		\includegraphics[width=.48\textwidth]{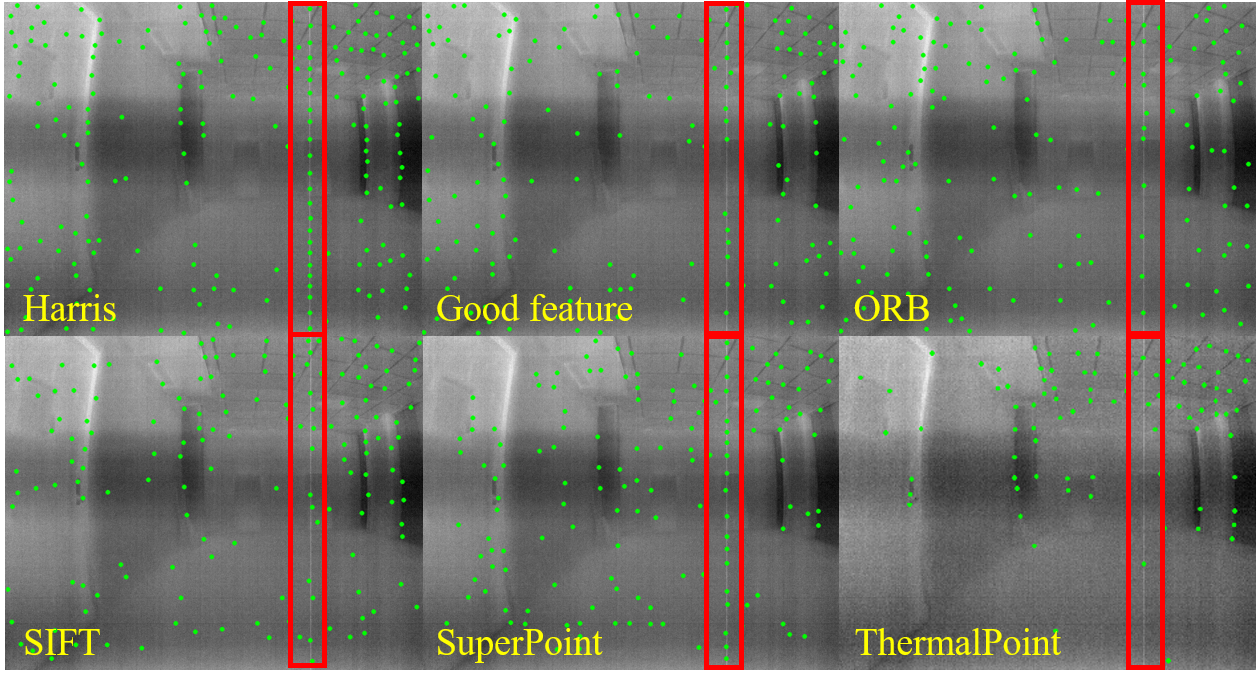} %1.png是图片文件的相对路径
		\caption{The anti-noise comparison of different feature detectors in the presence of fixed-pattern noise. The area marked by the red rectangle presents the effect of fixed-pattern noise, which appears as vertical stripes of different brightness in the thermal images.}
		\label{fig:feaure_noise_test}
	\end{figure}

	Fig.\ref{fig:feaure_illumi} shows the anti-photometric comparison of different feature detectors with sudden photometric changes. We can easily find that deep learning-based methods (ThermalPoint and SuperPoint) can robustly detect enough features even on the low contrast thermal images. However, the classical feature detectors like Harris, Good Feature, ORB and SIFT struggle in the presence of sudden photometric changes. Fig.\ref{fig:feaure_noise_test} shows the anti-noise comparison of different feature detectors in the presence of fixed-pattern noise appearing as vertical stripes. It can be seen that ThermalPoint extracts few features on the stripe while other methods detect many features on the stripe.

The reason behind this is that our deep-learning-based method can learn anti-photometric and anti-noise abilities through a training process. In contrast, most of the classic detectors are based on the gradient information of images; Therefore, it is difficult for classic detectors to overcome sudden photometric changes. 
	
	%\subsubsection{ Anti-noise Comparison}
	%Fig.\ref{fig:feaure_noise_test} shows the anti-noise comparison of different feature detectors in the presence of fixed-pattern noise. The area marked by the red rectangular presents the effect of fixed-pattern noise, which appears as vertical stripes of different brightness in the thermal images. We can easily find that Harris, Good feature, ORB, SIFT and SuperPoint  detect many features on the stripe. In comparison, our method extract few features on the stripe, which verifies that our algorithm has better anti-noise capability.
	
	%\begin{figure}[htbp]
	%	\centering
	%	\includegraphics[width=.4\textwidth ]{experiment_result/feature_illumi_test_jfr_mine/feature_illumi_jfr_mine.png} 
	%	\caption{The anti-photometric comparison of different feature detectors with sudden photometric changes}
	%	\label{fig:feaure_illumi}
	%\end{figure}
	
	\subsubsection{Quantitative Robustness Analysis} \label{Quantitative Analysis of feature detection }
	To further quantitatively analyze the robustness of feature detectors under viewpoint changes and photometric changes on thermal images, we use the repeatability score \cite{superpoint} to evaluate the robustness for each method. If the repeatability score is high, the corresponding method has a good ability of anti-photometric and anti-viewpoint changes. To test the repeatability on thermal images, we apply random homography transformations on our outdoor thermal image sequences to obtain corresponding thermal image pairs. We compute a maximum of 500 points for all systems at a 240x320 resolution. The Non-Maximum Suppression of all systems is 8 and the correctness threshold \cite{superpoint} $\varepsilon$ is 3. Table \ref{tab: feature detectors repeatability} presents the repeatability score for each method with all kinds of changes (viewpoint and photometric). It can be seen that ThermalPoint is the most repeatable under photometric and viewpoint changes.

	\begin{table}[h]
		\begin{center}
			\caption{Repeatability results of different feature detectors on our simulated dataset }
			
			\label{tab: feature detectors repeatability}
			
			\setlength{\tabcolsep}{10pt}
			\begin{tabular}{llllllllllllp{1cm}}
				\toprule %顶部
				Approach  			& Repeatability            	     \\ 
				\midrule
				Harris              & 0.4313                         \\ 
				Good feature  		& 0.6253   	                     \\
				ORB  				& 0.5748                 	     \\
				SIFT  				& 0.4976   	                     \\
				SuperPoint  		& 0.7166   	                     \\
				Our method          &\textbf{0.7674}                \\
				\midrule %低部线
			\end{tabular}
			\vspace{-8mm}
		\end{center}
	\end{table}
	
	\subsection{Robustness and Run-time Comparison of ThermalPoint Tracking}
	\label{Analysis of feature tracking }
\begin{figure}[b]
		\centering
		\includegraphics[width=.48\textwidth]{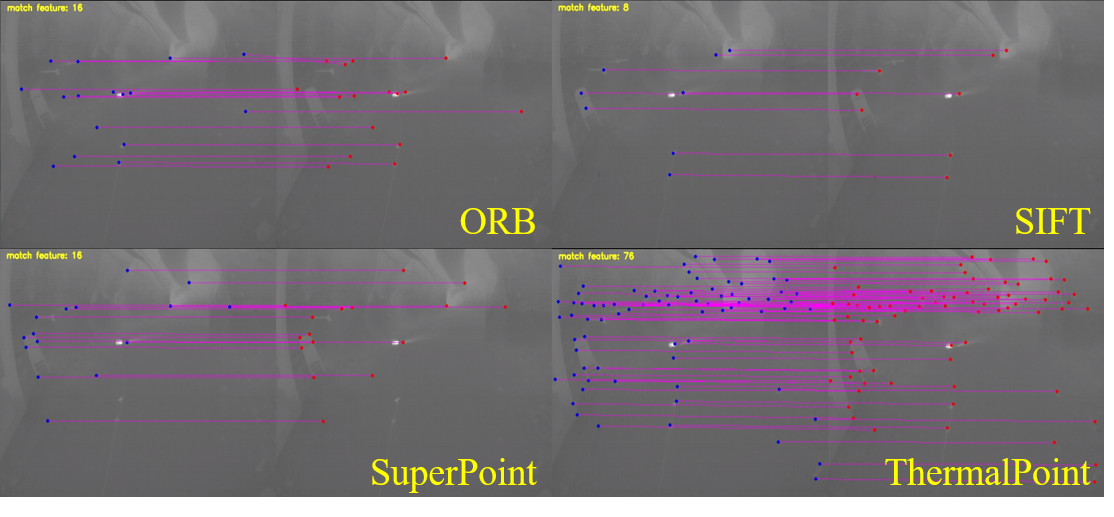}
		\caption{Samples of different feature tracking method in underground mine dataset.}
		\label{fig:feaure_math_jfr}
	\end{figure}

\begin{figure}[h]
		\centering
		\includegraphics[width=.4\textwidth, height= 0.36 \columnwidth ]{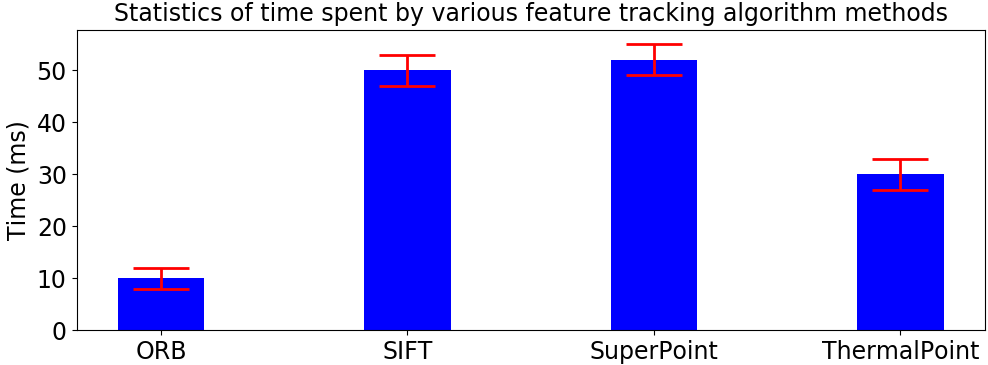} %1.png是图片文件的相对路径
		\caption{Average time for
detecting and tracking features between two frames}
		\label{fig:feaure_track_time_statistic}
	\end{figure}

	% \todo{提供一个ATE表格，衡量特征点在红外图像追踪成功的次数 （或者参考 GCNV2的写法）}
	To verify the robustness of ThermalPoint tracking, we test our method in an underground mine environment shown in Fig. \ref{fig:feaure_math_jfr}. During this test, we track a maximum of 160 points for all systems at a 640x512 resolution for every frame. The Non-Maximum Suppression (NMS) of all systems is 8. We find that ThermalPoint can track significantly more points than other algorithms on thermal images with low contrast. Also, we evaluated the real-time performance of various feature tracking algorithms (Fig. \ref{fig:feaure_track_time_statistic}). Our feature tracking algorithm takes significantly less time than SuperPoint.
	
	%\begin{figure}[htbp]
	%	\centering
	%	\includegraphics[width=.4\textwidth ]{experiment_result/feature_match/cmu_feature_math.png} %1.png是图片文件的相对路径
	%	\caption{ Sample of different feature tracking method in indoor dataset. The purple lines show correct correspondences. }
	%	\label{fig:feaure_math_cmu}
	%\end{figure}  

	\subsection{Accuracy Comparison of State Estimation in Indoor Environments} 
	
	\begin{figure}[htbp]
		\centering
		\includegraphics[width=.48\textwidth ]{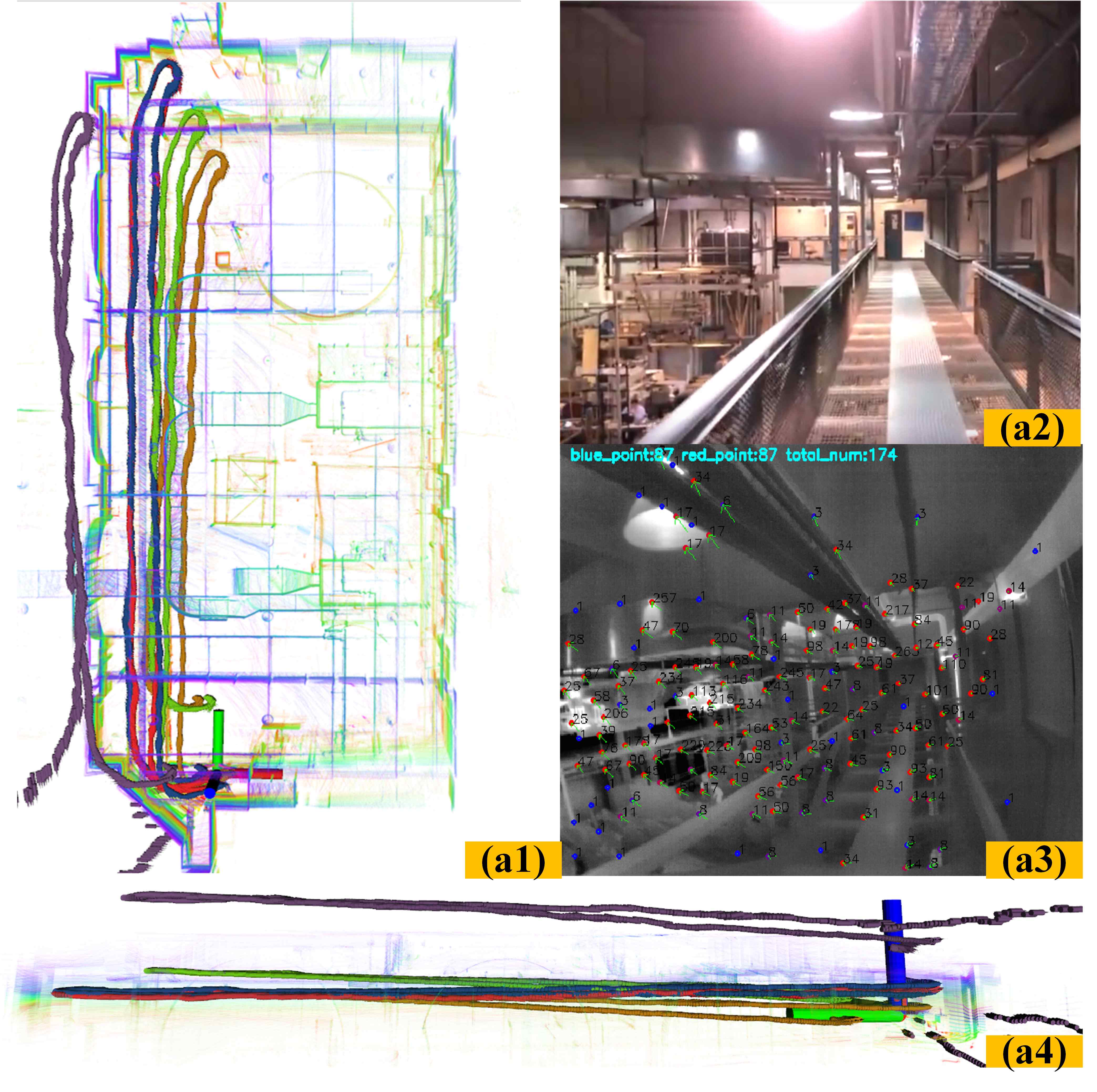} 
		\caption{(a1) and (a4) show the accuracy comparison of estimated trajectories of Ours (blue line), VINS (green line), OKVIS (yellow line) and R-VIO (purple line) against ground-truth (red line) in our laboratory environment as shown in (a2). DSO trajectory was excluded from the plots due to multiple failures. (a3) shows the quality of thermal images and  the feature detection result of our method in that scene.}
		\label{fig:accuracy}
	\end{figure}
To provide a comprehensive accuracy evaluation, the proposed method was compared with four state-of-the-art visual odometry methods, namely VINS \cite{VINS}, OKVIS [25], R-VIO [23] and DSO [19] on thermal imagery. The selections of the above methods are made mainly for two reasons: 1) Most of the thermal odometry approaches \cite{mouats2014multispectral,poujol2016visible,dai2019multi} require both visual and thermal images to achieve good motion estimation. However, our method works purely on thermal imagery; 2) Since most of the above methods use classic feature detection techniques, by comparing these methods, the effect of feature detection on thermal odometry will be clearly presented.
	
	To achieve a more fair comparison for state-of-the-art methods, we choose our laboratory environment with more heated sources (Fig. \ref{fig:accuracy}(a2)) and obtain high contrast thermal images (Fig. \ref{fig:accuracy}(a3)), which will be easier for state-of-the-art methods to achieve feature detection and tracking on thermal images. We held our device and walked 85 m along the bridge (Fig. \ref{fig:accuracy}(a2)) and returned back to the same position including purely rotational motion. The ground-truth is  provided by LOAM \cite{loam} algorithm.
	
	Fig. \ref{fig:accuracy}(a1,a4) shows the accuracy comparison of estimated trajectories in indoor environments. We can easily find that the trajectory provided by our method (blue line) is the closest to the ground-truth (red line), which suggests that our method is more accurate than other methods. 
	%Fig..\ref{fig:accuracy_xyz} and Fig..\ref{fig:accuracy_rpy} show the comparison results of derived translation and orientation  for each axis.
	The Absolute Trajectory Error (ATE)  and RPE \cite{benchmark} are presented in Table \ref{tab:accuracy}. It can be noted that the overall ATE and RPE errors of our method are very small.
	
	%\begin{figure}[htbp]
	%	\centering
	%	\includegraphics[width=0.5\textwidth]{experiment_result/odometry_comparison/xyz.png} 
	%	\caption{The translation comparison of Ours, VINS, OKVIS and R-VIO along each axis. DSO trajectory was excluded because of multiple failures}
	%	\label{fig:accuracy_xyz}
	%	\vspace{-8mm}
	%\end{figure}
	%
	%\begin{figure}[htbp]
	%	\centering
	%	\includegraphics[width=0.5\textwidth]{experiment_result/odometry_comparison/rpy.png} 
	%	\caption{The rotation comparison of Ours, VINS, OKVIS and R-VIO along each axis.DSO trajectory was excluded from the plots due to multiple failures.}
	%	\label{fig:accuracy_rpy}
	%\end{figure}
	
	\begin{table}[h]
		\begin{center}
			\caption{Accuracy evaluations of OURS, VINS,OKVIS,R-VIO and DSO methods operating on thermal images in indoor environments}
			\setlength{\tabcolsep}{10pt}
			\newcommand{\tabincell}[2]{\begin{tabular}{@{}#1@{}}#2\end{tabular}}
			\begin{tabular}{llllllllllllp{1cm}}
				\toprule
				\multirow{2}{*}{} 						& \multicolumn{2}{c}{ATE (in m)} 	& \multicolumn{2}{c}{RPE (in m)}    	     \\ \midrule
				Approach  					  			& RMSE     	& MAX     	& RMSE          &  MAX    \\ \midrule
				OURS                				& \textbf{0.359}   	&\textbf{0.866}    & \textbf{0.007}  &  \textbf{0.153}     \\
				VINS  					& 2.406   	& 3.902   & {0.033}   &  0.165         \\
				OKVIS  					& 3.199   	& 5.517   & {0.011}   &  0.258        \\
				R-VIO  					& 3.502   	& 4.637   & {0.029}   & 0.169          \\
				DSO  					& fail   	& fail   & fail   & fail   \\
				\midrule
			\end{tabular}
			\label{tab:accuracy}
			\vspace{-4mm}
		\end{center}
	\end{table}
	
	%\begin{figure}[htbp]
	%	\subfigure[light fog scene]{
	%		\label{scene1}
	%		\includegraphics[width=0.45\columnwidth]{experiment_result/fog_scene_pics/light_fog_pic.png}
	%	}
	%	\subfigure[heavy fog scene]{
	%		\label{scene2}
	%		\includegraphics[width=0.45\columnwidth]{experiment_result/fog_scene_pics/heavy_fog_pic.png}
	%	}
	%	\caption{The real scene of highway}
	%	\label{fig:realscene}
	%	\vspace{-5mm}
	%\end{figure}

	\subsection{Robustness Comparison of Pose Estimation in Smoke Environments } 
To evaluate the robustness of our method in challenging environments, we set up smokey environments with a smoke generator. 
We compare our method with the famous LOAM \cite{loam} algorithm to evaluate the accuracy and robustness.
The reason for selecting LOAM algorithm is that it can achieve robust pose estimation in light smoke environments while visual odometry will fail in such environments. 

%Secondly, 
% % LOAM algorithm is famous for its robustness and accuracy on pose estimation. Therefore, comparing with the LOAM algorithm is more persuasive to illustrate the advantages of thermal odometry.

    In terms of testing the accuracy of trajectories, we use a pre-built point cloud map as a ground-truth map to present the estimated trajectory. If the estimated trajectory is accurate, it will coincide with the map well. In this way, we can qualitatively evaluate the accuracy of the trajectory.
    In terms of the robustness, we use a smoke machine to adjust the amount of smoke and build light smoke environments (Fig. \ref{fig:first_page}(a)) and heavy smoke environments (Fig. \ref{fig:first_page}(b)) to evaluate the performance of the proposed method. In real scenes, we carried our handheld device and walked from room A to the smoke room by the hallway and returned back to the same position in room A as shown in Fig. \ref{fig:pose_lightsmoke}(a).

	\subsubsection{State estimation comparison under light smoke environments}

	\begin{figure}[htbp]
		\centering
		\includegraphics[width=0.48\textwidth]{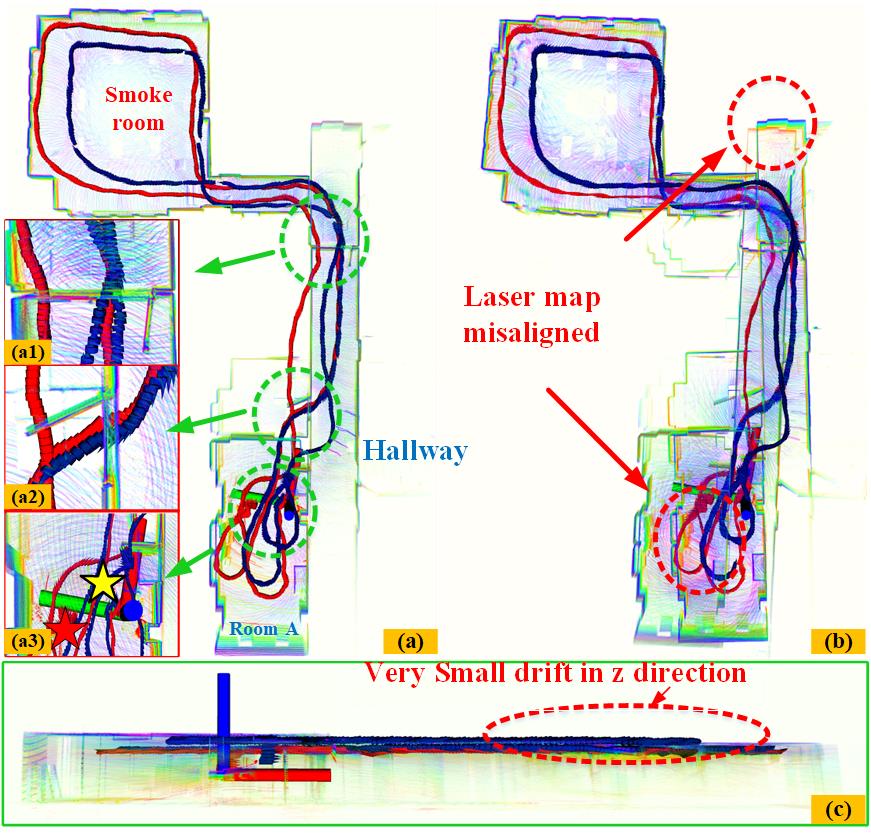} 
		\caption{The state comparison of our method and LOAM under light smoke environments. (a) is ground-truth map and the red and blue trajectories are generated by LOAM and our algorithm respectively. The green circles show more detailed inspections (a1-a3) on the trajectories and map to illustrate the accuracy. (b) shows the actual point cloud  misalignment from LOAM. (c) shows estimated trajectories in the pre-built map in z direction.}
		\label{fig:pose_lightsmoke}
	\end{figure}
	
	In this section, we present the trajectory comparison result between our algorithm (blue line) and LOAM \cite{loam} (red line) in light smoke environments.
	% The map shown in Fig.\ref{fig:pose_lightsmoke}(a) is ground-truth map. 
	%We carried our handheld device and walked from room A to the  light smoke room as shown in Fig.\ref{fig:first_page}(a) and return back to the same position in room A. 
	% In Fig.\ref{fig:pose_lightsmoke}(a), the red trajectory is generated by LOAM and the blue trajectory is generated by our algorithm. The green circles show more detailed inspections on the trajectories and map to illustrate the accuracy.
Fig. \ref{fig:pose_lightsmoke}(a1,a2) shows the trajectories crossing a narrow doorway. Our algorithm (blue trajectory) can pass through the doorway while Lidar trajectories fail and collide with the wall, meaning that Lidar odometry has a relatively large drift. It's worth noting that the narrow doorway has a width of only 0.67 meters.
	% Fig. \ref{fig:pose_lightsmoke}(a2) shows another case of passing through narrow doorway during the return trajectory, and we can see a similar result of our method (blue trajectory) being able to pass through the narrow doorway while lidar odometry (red trajectory) fails and has large drift.
Fig. \ref{fig:pose_lightsmoke}(a3) shows the result of both trajectories returning to the origin(xyz axes). The red star and yellow star in the figure represent the final position of Lidar odometry and our algorithm, respectively. As we can see, our algorithm's final position is closer to the original start position(xyz axes), indicating that our algorithm is more accurate in this experiment.
	%The whole trajectory has a length of 106 meters, and the drift of final position of our algorithm is only 0.2 meters.
Fig. \ref{fig:pose_lightsmoke}(b) shows the actual point cloud misalignment from LOAM under light smoke. This suggests that the Lidar point cloud map has a large drift in such environments. Fig. \ref{fig:pose_lightsmoke}(c) shows the trajectories in the pre-built map in the z-direction, and it can be seen that our algorithm has a small z-directional drift.
	
	\begin{figure}[htbp]
		\centering
		\includegraphics[width=0.48\textwidth]{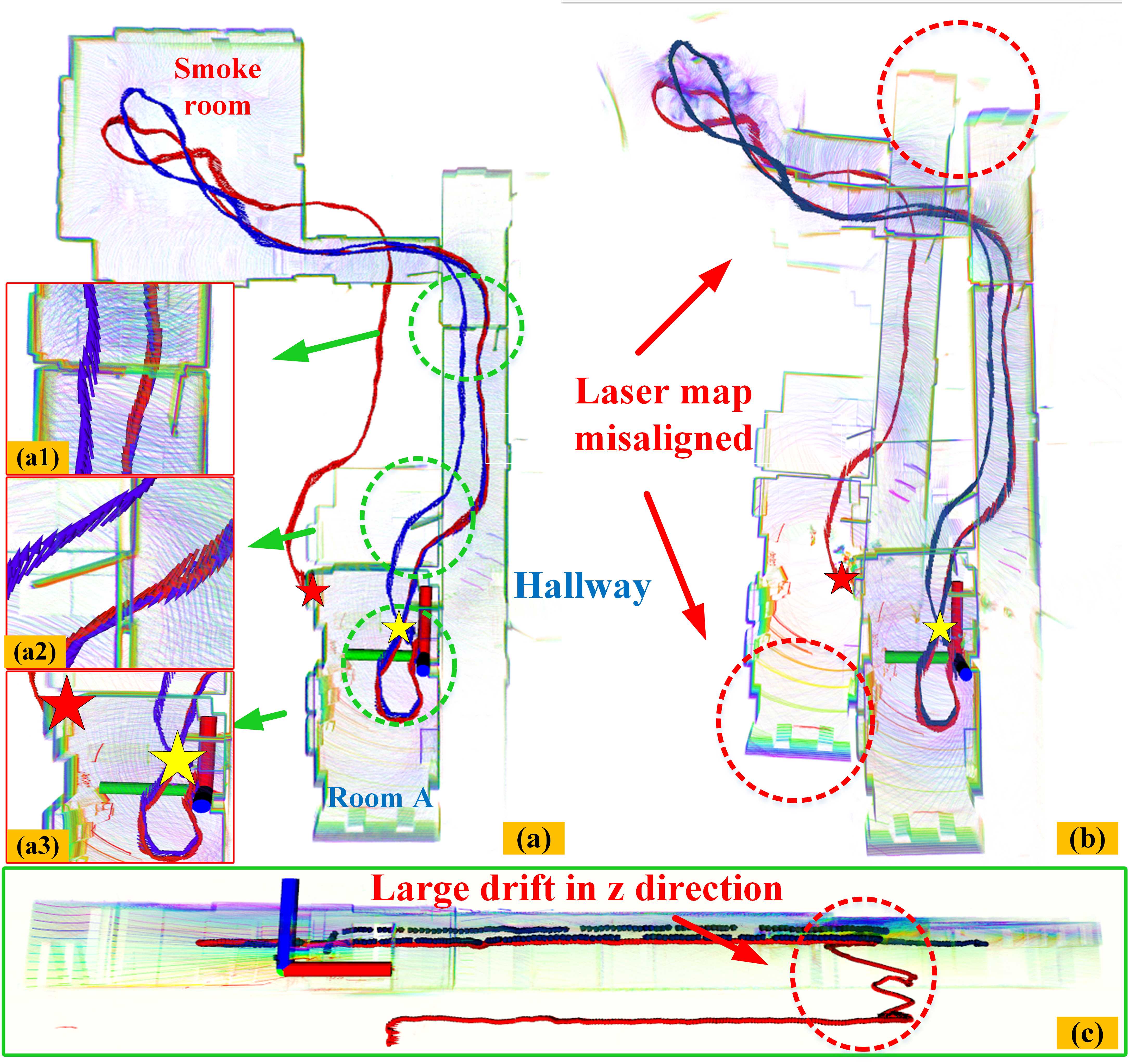} 
		\caption{The state comparison of our method and LOAM under heavy smoke environments. (a) is ground-truth map and the red and blue trajectories are generated by LOAM and our algorithm respectively. The green circles show more detailed inspections (a1-a3) on the trajectories and map to illustrate the accuracy. (b) shows the actual point cloud misalignment from LOAM. (c) shows estimated trajectories in the pre-built map in z-direction.}
		\label{fig:pose_heavysmoke}
	\end{figure}
	
	\subsubsection{State Estimation Comparison under Heavy Smoke Environments}
	
	To further test the robustness of our algorithm, we increase the amount of smoke, as shown in Fig. \ref{fig:first_page}(b). 
	%The point cloud map in Fig. \ref{fig:pose_heavysmoke}(a) is ground-truth map.
	% The point cloud map in figure \ref{fig:pose_heavysmoke}(a) is built with no fog in the environment and is used as ground truth map.
	%Similar to the previous experiment with light smoke, we carried our handheld device to travel from room A to the smoke room through a long corridor, and then returned to the same position in room A.
	%   In Fig.\ref{fig:pose_heavysmoke}(a), red trajectory represents the Lidar odometry and blue trajectory represents our algorithm. 
Fig. \ref{fig:pose_heavysmoke}(a1,a2) shows the details of trajectories passing through the narrow doorway, and we can see that our odometry (blue line) can pass through the door while the Lidar odometry (red line) fails and collides with the wall. Fig. \ref{fig:pose_heavysmoke}(a3) shows the result of trajectories returning to the origin (xyz axes), where the red and yellow star represent the final position of Lidar odometry and our algorithm respectively. We can see that the final position of our algorithm is closer to the origin, indicating that our trajectory is more accurate in the heavy smoke environment. 
	Fig. \ref{fig:pose_heavysmoke}(b) shows the actual point cloud  misalignment from LOAM \cite{loam} under heavy smoke. This means that Lidar point cloud map has a large drift in such environments. Fig. \ref{fig:pose_heavysmoke}(c) shows the trajectories in the pre-built map in z-direction, and it can be seen that our algorithm has small z-directional drift while Lidar odometry (red trajectory) has large drift. This experiment shows that our algorithm is more robust under heavy smoke.

	%\begin{table*}[htbp]
	%	\begin{center}
	%		\caption{Accuracy evaluations of LOAM, SuMa and ours method on KITTI datasets (The blue numbers highlight the best result for each specific dataset among all the compared algorithms)}
	%		\setlength{\tabcolsep}{10pt}
	%		\newcommand{\tabincell}[2]{\begin{tabular}{@{}#1@{}}#2\end{tabular}}
	%		\begin{tabular}{llllllllllllp{1cm}}
	%			\toprule
	%			\multirow{2}{*}{} 	& \multicolumn{10}{c}{Relative errors averaged over trajectories of 100m to 800m length: relative translation error in $\%$} \\ \midrule
	%			Approach  					  			& 00     	& 01     	& 03     & 04     	& 05      	& 06     &07  &09  &10 	& Average  \\ \midrule
	%			OURS(Frame-to-Frame)                			& 1.3   	& 3.4       & 1.4    & 1.2     	& 1.6    	&1.3     &2.1 &2.3 &1.2 &1.75 \\
	%			OURS(Frame-to-Model)  					    & 0.9   	& \textcolor{blue}{0.8}   &\textcolor{blue}{0.9}   & \textcolor{blue}{0.6} &0.9 &0.6 &1.2   &\textcolor{blue}{0.8}       & \textcolor{blue}{0.8} 	& 0.83 \\
	%			LOAM  					& 0.8   	& 1.4   & \textcolor{blue}{0.9}   & 0.7   &\textcolor{blue}{0.6}       & 0.7 &\textcolor{blue}{0.6} &\textcolor{blue}{0.8} & \textcolor{blue}{0.8}& \textcolor{blue}{0.81}\\
	%			SuMa (Frame-to-Modle)  					& \textcolor{blue}{0.7}   	& 1.7   & 1.0   & \textcolor{blue}{0.6}   &\textcolor{blue}{0.6}       &\textcolor{blue}{0.5} 	& 0.7 &0.9 &0.9 &0.88\\
	%			\midrule
	%		\end{tabular}
	%		\label{tab:kitti}
	%	\end{center}
	%\end{table*}
	%\vspace{-3mm}
	\section{Conclusion} 
	\label{sec:conclusion}
	In this paper, we propose "ThermalPoint", a CNN-based feature detection network specifically tailored for producing keypoints on thermal images. To achieve real-time performance, we combine ThermalPoint with a novel radiometric feature tracking method and establish reliable correspondences between sequential frames. Finally, a deep feature based thermal-inertial odometry (TP-TIO) estimation framework is proposed and evaluated thoroughly in various visually-degraded environments. Experiments show that our method outperforms state-of-the-art visual and laser odometry methods in smoke-filled environments and achieves competitive accuracy in normal environments.
    In the future, we will add a loop closures module through ThermalPoint and correct accumulated drift for long term navigation.

	%%%%%%%%%%%%%%%%%%%%%%%%%%%%%%%%%%%%%%%%%%%%%%%%%%%%%%%%%%%%%%%%%%%%%%%%%%%%%%%% 
	\section{ACKNOWLEDGMENT}
This work was partially supported by DARPA agreement \#HR00111820044. Also, this work was supported by National Natural Science Foundation of China (No.61573091), the Fundamental Research Funds for the Central Universities (No.N182608003, No.N172608005), Natural Science Foundation of Liaoning Province (20180520006) and Major Special Science and Technology Project of Liaoning Province (No.2019JH1/10100026).
We also wish to thank Dr. Shehryar Khattak for his constructive advice.
	
	%and Open Research Project of the State Key Laboratory of Industrial Control Technology, Zhejiang University,China(No.ICT170302). 
	\balance
	
	\bibliographystyle{IEEEtran}
	\bibliography{mybibfile}

	%\begin{table}[h]
	%\caption{An Example of a Table}
	%\label{table_example}
	%\begin{center}
	%\begin{tabular}{|c||c|}
	%\hline
	%One & Two\\
	%\hline
	%Three & Four\\
	%\hline
	%\end{tabular}
	%\end{center}
	%\end{table}

	%   \begin{figure}[thpb]
	%      \centering
	%      \framebox{\parbox{3in}{We suggest that you use a text box to insert a graphic (which is ideally a 300 dpi TIFF or EPS file, with all fonts embedded) because, in an document, this method is somewhat more stable than directly inserting a picture.
	%}}
	%      %\includegraphics[scale=1.0]{figurefile}
	%      \caption{Inductance of oscillation winding on amorphous
	%       magnetic core versus DC bias magnetic field}
	%      \label{figurelabel}
	%   \end{figure}
	
\end{document}